\documentclass[sn-nature]{sn-jnl} 

\usepackage{graphicx}%
\usepackage{multirow}%
\usepackage{amsmath,amssymb,amsfonts}%
\usepackage{amsthm}
\usepackage{mathrsfs}%
\usepackage[title]{appendix}%
\usepackage{xcolor}%
\usepackage{textcomp}%
\usepackage{manyfoot}%
\usepackage{booktabs}%
\usepackage{algorithm}%
\usepackage{algorithmicx}%
\usepackage{algpseudocode}%
\usepackage{listings}%
\usepackage{geometry}
\usepackage{pdflscape}	
\usepackage{array}
\usepackage[textsize=tiny]{todonotes}

\usepackage{nameref}
\usepackage{chngcntr} 
\usepackage{marginnote}
\usepackage{etoolbox}
\newtoggle{lmargin}
\newcommand{\alternatingtodo}[2][]{%
\iftoggle{lmargin}%
{%
\todo[#1]{#2}
\togglefalse{lmargin}
}{%
{%
\let\marginpar\marginnote%
\reversemarginpar%
\todo[#1]{#2}%
}%
\toggletrue{lmargin}%
}%
}%
\usepackage{graphicx} 
\graphicspath{ {figures/} }
\DeclareUnicodeCharacter{03B2}{\ensuremath{\beta}}
\begin{document}
\title[Article Title]{Indication Finding: a novel use case for representation learning}

\author{Maren Eckhoff$^1$,
Valmir Selimi$^1$,
Alexander Aranovitch$^1$,
Ian Lyons$^1$,
Emily Briggs$^1$,
Jennifer Hou$^1$,
Alex Devereson$^1$,
Matej Macak$^1$,
David Champagne$^1$,
Chris Anagnostopoulos$^1$\\
\\ 
$^1$QuantumBlack, AI by McKinsey}

\abstract{Many therapies are effective in treating multiple diseases. We present an approach that leverages methods developed in natural language processing and real-world data to prioritize potential, new indications for a mechanism of action (MoA). We specifically use representation learning to generate embeddings of indications and prioritize them based on their proximity to the indications with the strongest available evidence for the MoA. We demonstrate the successful deployment of our approach for anti-IL-17A using embeddings generated with SPPMI and present an evaluation framework to determine the quality of indication finding results and the derived embeddings.}

\keywords{representation learning, SPPMI, real-world evidence, real-world data, drug repurposing, indication finding}

\maketitle


\section{Introduction}\label{introduction}

Prioritizing which indications to pursue is crucial to the drug development process. 
This applies to both treatments with a novel mechanism of action, yet to enter the market, and those that are already approved for multiple indications, requiring life cycle management. 
For life cycle management, Krishnamurthy et al. identified that 30–40\% of drugs approved between 2007 and 2009 have been repurposed for a new indication.
However, reports indicate that only about 30\% of repurposing efforts are successful\,\cite{krishnamurthy_drug_2022}. 
These outcomes can be a consequence of the indications that have been selected for exploration. 
Krishnamurthy et al.\ found that about 78\% of abandoned repurposing attempts are due to low clinical efficacy\,\cite{krishnamurthy_drug_2022}.
Traditionally, drug repurposing has been based on guidance from key opinion leaders and insights from -omics and literature data\,\cite{liu_silico_2013, kulkarni_drug_2023}. 
Other approaches analyse spontaneous use (i.e., clinician-led use in unlicensed indications) \cite{wittich_ten_2012} or study effectiveness for comorbid diseases (i.e., observed through incidental exposure) \cite{wang_disease_2018, lin_integrating_2023, kasznicki_metformin_2014}.
Case reports or epidemiological studies are increasingly considered in the regulatory process, however, they can scan only a very small set of indications and patient journeys, and are naturally limited to marketed MoAs with significant use.

Computational approaches leveraging real-world data (RWD) for hypothesis generation to improve success in drug repurposing efforts are an active area of research\,\cite{tan_drug_2023}. 
For example, Kim et al.\ identify associations between changes in laboratory results (indicative of disease progression) and prescription of therapeutic agents to hypothesize new treatments for diseases \cite{kim_high-throughput_2020}. 
Cosmin et al.\ have used retrospective cohort analysis to surface repurposing opportunities for COVID-19 \cite{cosmin_a_bejan_drugwas_2021}. 
Drug repurposing using knowledge graphs that encode relationships between real world entities is another active area of research \cite{li_drug-cov_2023, ma_kgml-xdtd_2023}. 
Unlike the first two approaches, our approach does not require the treatment of interest or other treatments with the same MoA to be prescribed. 
It also does not rely on lab test results which are often scarce or unavailable in RWD sources.

Recent breakthroughs in artificial intelligence have been facilitated by the application of self-supervised learning to massive datasets \cite{brown_language_2020, touvron_llama_2023} to create models that speak the language of the domain of interest.
One way to access the information learned is through the models' internal representations of units, often called embeddings. 
In RWD, the units might be clinical events such as diagnoses, prescriptions, and procedures, and the embeddings capture the phenotypic similarity between these events \cite{jiang_health_2023, wornow_shaky_2023, chen_disease_2021}. 
Recently, Hong et al. utilized representations generated from RWD to identify diagnoses closely associated with a given prescription, demonstrating broader applications of embeddings in the context of RWD are possible \cite{hong_clinical_2021}.

In this paper we present a novel approach to indication finding providing broad quantitative prioritisation for established as well as novel MoAs. 
The output can inform research efforts and clinical development plans.
The approach relies on representation learning to obtain embeddings of clinical events codified in RWD. 
For a given MoA, referential diseases that the MoA is likely to be effective in treating based on orthogonal sources are identified.
We then prioritize indications that are close to the referentials in the embedding space. 
The assumption is that diseases which are similar to a set of distinct referentials share the pathway and underlying pathology on which the studied MoA acts. 
Finally, we perform a quantitative evaluation and ranking of indications at scale.

Many methods to generate embeddings on RWD have been developed \cite{wornow_shaky_2023}. 
In this work, we use SPPMI to derive embeddings \cite{levy_neural_2014}. 
SPPMI is a well-established method that is similar to the popular word2vec embedding method \cite{mikolov_efficient_2013}, however, the embeddings can be computed directly and are therefore significantly more resource efficient on large datasets making the approach more widely accessible.

To assess the quality of embeddings and demonstrate the clinical validity of our approach, we developed an evaluation framework that provides qualitative and quantitative metrics. 

To summarize, the main contributions of this paper are:
\begin{enumerate}
  \item Introduction of a new approach to indication finding leveraging RWD and representation learning
  \item Definition of an evaluation framework to determine the quality of clinical event representations and indication ranking
  \item Application of the new approach and evaluation in a case study for the anti-IL-17A MoA

\end{enumerate}

\section{Results}\label{results}

In this section, we introduce the new indication finding approach and show that it ranks diagnoses for which efficacy has been demonstrated in clinical trials higher than those for which clinical trials have failed to demonstrate efficacy.

\subsection{Introduction to the approach}\label{introduction_to_the_approach}

\begin{figure}[h]
\includegraphics[width=1\textwidth]{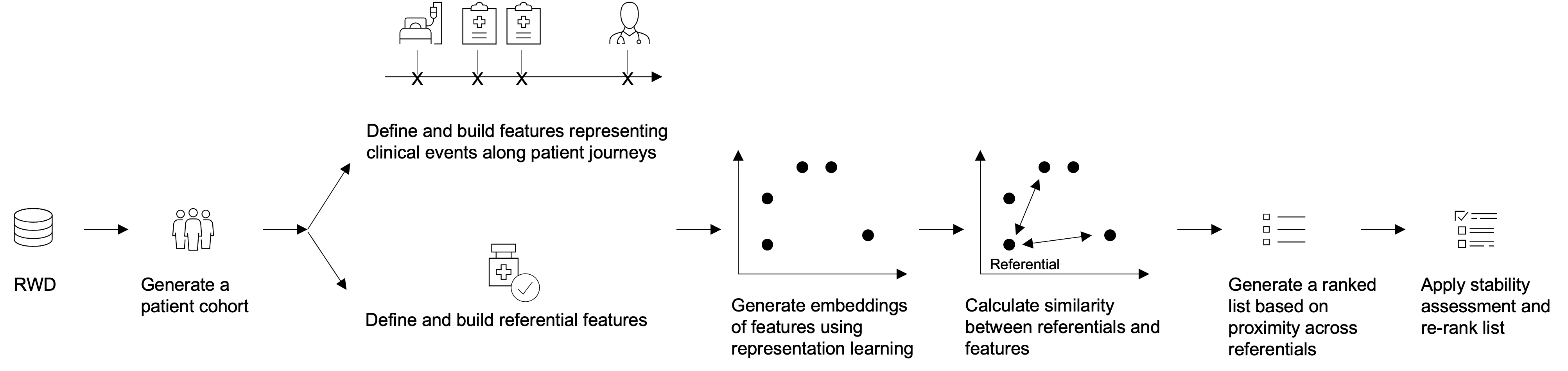}
\centering
\caption{Overview of the indication finding approach.}
\label{fig:diagram}
\end{figure}

The basis for the analysis is a large cohort of patients from a real world data source with links to the MoA of interest. 
Patient journey characteristics like diagnoses, prescriptions, procedures and demographics are defined as features along the longitudinal patient records. 
In addition, indications for which there is the highest confidence that the studied MoA can be effective in treating are selected, and referred to as \textit{referentials} (Section \ref{feature_definition}).
For established MoAs these referentials may be based on approved indications or  indications with success in phase II / III clinical trials. For novel MoAs, referentials may be derived from earlier trial data, -omics, literature or molecular knowledge graphs. 
For the anti-IL-17A case study, indications approved for treatment with the MoA were selected: ankylosing spondylitis, plaque psoriasis and psoriatic arthritis (NCT numbers see Table \ref{table: ref_and_control_icd_codes}). 
With the cohort and features defined, embeddings of the features are learned (Section \ref{embedding_approach}). Potential indications are scored based on the cosine similarity of their embeddings to the embeddings of referentials, and ranked across the referentials. 
Finally, the stability of the ranking is assessed and unstable indications are removed (Section \ref{ranked_list_of_indications}). The full workflow is illustrated in Figure\,\ref{fig:diagram}.

\subsection{Evaluation Framework}\label{evaluation_framework}

Indication finding is a semi-supervised learning problem, as typically only a small number of indications have been tested.
Hence, particular effort is required to assess the quality of the results. The final ranking of prioritized indications was evaluated quantitatively using positive and negative validations (see Section\,\ref{quantitative_ranked_list_evaluation}) and qualitatively through review by a medical professional. These approaches are commonly used to evaluate computational approaches for drug repurposing \cite{jang_inferring_2016, koren_machine_2018, vanhaelen_machine-learning-based_2019, paik_repurpose_2015, prinz_novel_2018}.

In addition, the quality of the learned representations was evaluated to build confidence in the evidence underpinning the ranked list. 
We defined two metrics to assess if the embeddings capture clinically meaningful information.
Firstly, we tested the ability to predict clinical events using embedding-based features, similar to Ruan et al. \cite{ruan_representation_2019}. 
Secondly, the interpretability of the learned embeddings is assessed by projecting into 2D and reviewing whether the spatial relationships are clinically meaningful. A similar method was described by Rupp et al., who noted that when patient embeddings were plotted in two dimensions, clusters of patients with related diseases appeared in close proximity to each other\,\cite{rupp_exbehrt_2023}.

\subsection{Evaluation of Indication Finding performance}\label{evaluation_ranked_indication_lists}
In the following sections, we review the validity of the ranked list leveraging orthogonal sources. 
This review is done by assessing the ranking of positive and negative validations and assessing the relevance of increased IL-17A in the pathogenesis of indications based on published evidence.

\subsubsection{Quantitative evaluation of ranked list}\label{quantitative_ranked_list_evaluation}
Evidence from clinical trials is commonly used to validate results from computational drug repurposing efforts \cite{jang_inferring_2016,koren_machine_2018}.
We defined \emph{positive} and \emph{negative} validations as diagnoses for which the MoA has been demonstrated to be effective or for which it has been shown to have inadequate efficacy, respectively.
Their absolute and relative ranking in the derived list was used to evaluate the approach.
If positive validations rank higher than negative validations, it indicates that the approach can provide value in differentiating indications with a high or low probability of success.

We selected as positive validations all diagnoses, defined by specific ICD-10 codes, for which the MoA has been shown to be effective in treating in Phase 2 or Phase 3 trials. 
These positive validations were required to have a specific ICD-10 code available and to be present in over 100 patients in the cohort. 
We ensured that the evidence produced by the trials did not affect the management of subpopulations with a validation diagnosis by assessing the prescription of anti-IL-17A over time; we excluded a diagnosis as a validation if there was an absolute increase in prescription of anti-IL-17A following publication of the clinical trial results to ensure our analysis was `blinded' to the evidence, meaning that the data on which analyses were performed did not have a direct link between specific assets and indications.
This resulted in exclusion of \emph{pityriasis rubra piliaris}. 
We also excluded validations that could be considered a sub-category of a referential indication; namely \emph{palmoplantar psoriasis} and \emph{generalized pustular psoriasis}, which are both subtypes of \textit{Psoriasis}. 
For anti-IL-17A, the five positive validations used were \emph{rheumatoid arthritis}, \emph{rosacea}, \emph{hidradenitis suppurativa}, \emph{non-infectious uveitis} and \emph{giant cell arteritis} \cite{kumar_exploratory_2020,blanco_secukinumab_2017,kimball_secukinumab_2023,letko_efficacy_2015,venhoff_efficacy_2021}.

Conversely, negative validations were defined as diagnoses for which the MoA failed to demonstrate efficacy in phase II/III clinical trials. Negative validations were required to satisfy the same inclusion criteria as positive validations. The indication \emph{dry eye} was excluded from negative validations due to lack of specificity to the anti-IL-17A MoA. For anti-IL-17A, the five negative validations were \emph{Crohn's disease}, \emph{congenital ichthyosis}, \emph{atopic dermatitis}, \emph{bullous pemphigoid} and \emph{COVID-19} \cite{wolfgang_hueber_secukinumab_2012,lefferdink_secukinumab_2023,ungar_phase_2021,grosskreutz_dry_2015,mangold_aaron_ixekizumab_2020}.

\begin{figure}[h]
\includegraphics[width=0.8\textwidth]{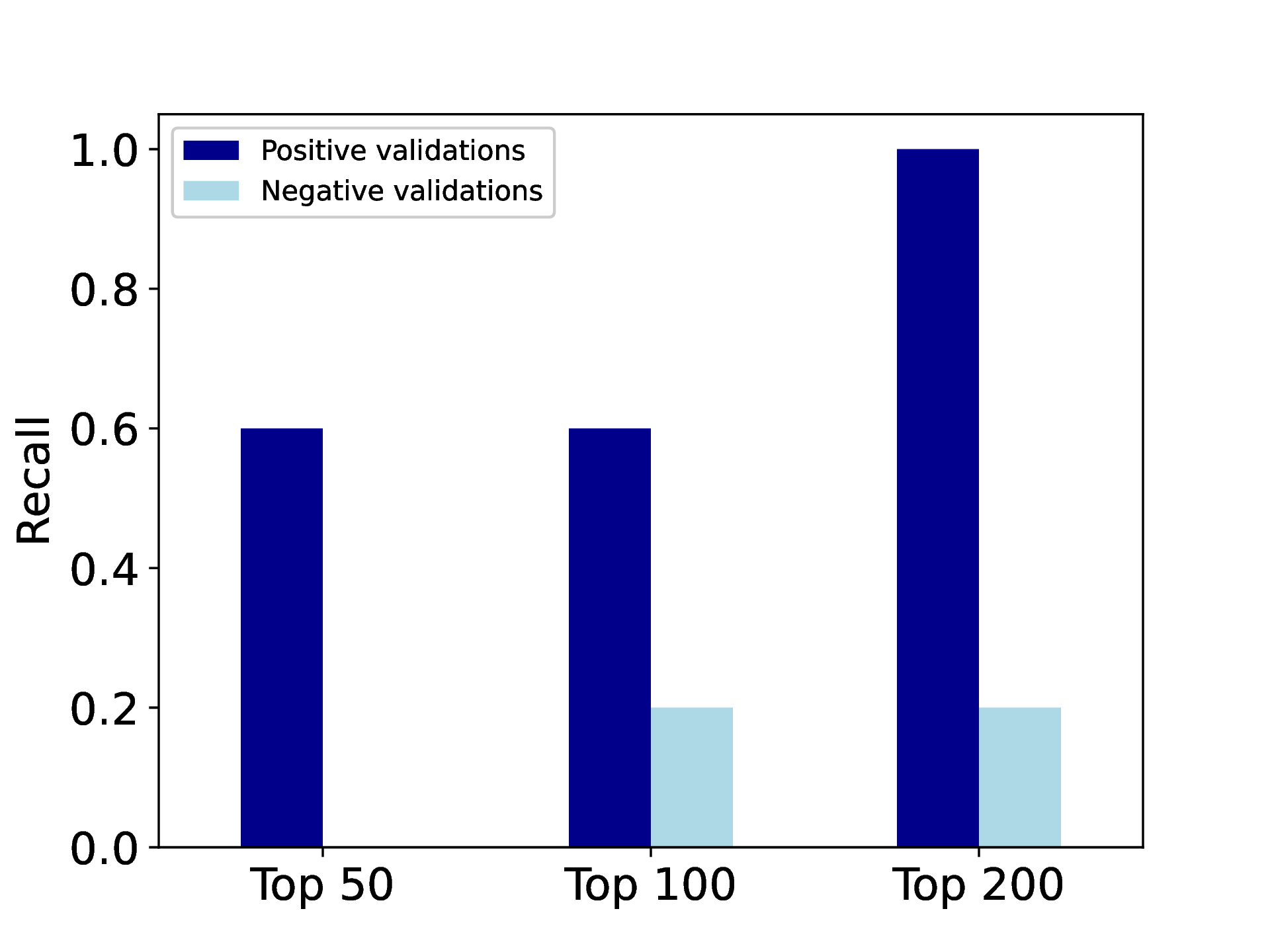}
\centering
\caption{Recall of positive validations and negative validations for the top 50, top 100, and top 200 ranked indications. There were five positive and five negative validations, respectively.}
\label{fig:pos_vs_neg_recall}
\end{figure}

Whilst both positive and negative validations have a biological link to the target of the MoA which led to their exploration in clinical trials, a successful indication finding approach ranks positive validations relatively higher, suggesting that it may be able to predict efficacy. 
In the anti-IL-17A case study, the top 50 indications contained 60\% of the positive validations and no negative validations. 
In the top 100, it was 60\% vs 20\%, and in the top 200, 100\% vs 20\% (see Figure \ref{fig:pos_vs_neg_recall} and Table \ref{table:evaluation_of_embeddings_table}). 
For comparison, selecting indications at random would yield about 6\%, 11\% and 23\% of positive validations within the top 50, 100 and 200, respectively. 

\subsubsection{Qualitative evaluation of ranked list}\label{qualitative_ranked_list_evaluation}
Literature review is commonly used to corroborate results of indication selection produced by computational approaches \cite{vanhaelen_machine-learning-based_2019, prinz_novel_2018}.
For the anti-IL-17A case study, an indication was deemed clinically plausible if a peer-reviewed publication could be identified demonstrating that increased IL-17A is involved in the pathology of the disease. 
The number of indications linked to high IL-17A in the top 100 indications was 81 (see Table \ref{table:evaluation_of_embeddings_table}).
Of the remaining 19, 11 were linked to downstream transcription factors or signalling proteins that are involved in the IL-17A pathway, therefore representing potentially novel targets for which there is yet to be an evidenced direct association with IL-17A.

\begin{table}[h]
\begin{tabular*}{\textwidth}{@{\extracolsep{\fill}}p{1cm}p{2cm}p{1.2cm}p{1.2cm}p{1cm}@{\hspace{0.6cm}}}
& & & & \textbf{Score} \\
\toprule%
\multirow{9}{2cm}[2.7em]{\textbf{Evaluation of Indication Finding performance}}&
\multirow{6}{3cm}[1.05em]{Quantitative review of ranked list (\% of validations in a given slice of the list)}&
\multirow{3}{1.2cm}[0.5em]{Positive validations}& Top 50 & 60.0\%\\\cmidrule{4-5}	
& & & Top 100 & 60.0\% \\\cmidrule{4-5}					
& & & Top 200 & 100.0\% \\\cmidrule{3-5}					
& & \multirow{3}{1.2cm}[0.5em]{Negative validations}& Top 50 & 0.0\% \\\cmidrule{4-5}	
& & & Top 100 & 20.0\% \\\cmidrule{4-5}	
& & & Top 200 & 20.0\% \\\cmidrule{3-5}	
\cmidrule{2-5}	
				
& \multirow{3}{4cm}{Qualitative review of the ranked list (\% of diagnoses that are clinically plausible in a given slice of the list)} 
& & Top 50	 & 96.0\%	\\\cmidrule{4-5}	
&& & Top 100	 &81.0\%		  \\\cmidrule{4-5}	
&& & Top 200	 &	60.5\%   \\\midrule
\multirow{3}{2cm}{\textbf{Evaluation of the quality of embeddings}}&\multirow{1}{4cm}{Clinical review of feature grouping in embedding space}& &\multirow{1}{4.5cm}{2D mapping of diagnosis embeddings demonstrates clinically explainable groupings (see Figure \ref{fig:embeddings})}\vspace{0.3cm}\\
& & & &  \\ 
& & & &    \\  \cmidrule{2-5}\\
&\multirow{1}{4cm}{Predictive task}& & &\textbf{AUC} \\\cmidrule{2-5}
& \multirow{1}{8cm}{Embedding features (75 dimensions)}& & &67.5\%  \\ 
&  \multirow{1}{8cm}{Baseline using original features (2972 dimensions)}& & &67.5\%  \\ 

 \bottomrule
\end{tabular*}
\caption{Overview of results for the anti-IL-17A case study. There was a total of five positive and five negative validations. We call an indication clinically plausible if there is at least one piece of evidence supporting the link along with the absence of substantial evidence to the contrary. Additional data is provided in Table \ref{tab:predictive_performance_by_task}}
\label{table:evaluation_of_embeddings_table}

\end{table}

\subsection{Evaluating the quality of embeddings for the anti-IL-17A MoA}\label{evaluation_of_embeddings}

In the following sections, we review the quality of the embeddings with regard to their performance in predictive tasks and clinical interpretability.

\subsubsection{Performance in prediction tasks}\label{predictive_performance}
We quantitatively evaluated the embeddings by assessing their predictive power as features in a variety of binary classification tasks.
The expectation is that good embeddings encode the information required to perform the tasks successfully on a similar level as the original clinical events while having only a fraction of the dimensionality. 
Each task was to predict the onset of a specific target disease one day ahead. 
We created six predictive tasks using target diagnoses of different prevalence and nature: stroke (prevalence: 6.1\%), myocardial infarction (5.4\%), neutropaenia (1.4\%), rheumatoid arthritis (11.2\%), chronic kidney disease (12.8\%) and type 2 diabetes mellitus (17.9\%).
Features used for the tasks were constructed by summing the embeddings in a six month look-back window from the index date. 
As a baseline model, we used occurrence counts as features. 
For each task, a cohort of 50,000 patients was obtained using sampling stratified on target.
Patients with the target diagnosis, who were observed less than six months before their first diagnosis, were discarded.
For patients without the disease, the index date was chosen at random.
Three classifiers were trained: logistic regression, random forest (RF) and a multi-layer perceptron (MLP), using Scikit-learn \cite{pedregosa_scikit-learn_2011} with default parameters. As a performance metric for each task, we used the greatest AUC across the three classifiers, evaluated on a holdout set of 10,000 patients. 
It should be noted that embeddings were not fine-tuned to the predictive tasks to assess the exact representations used for indication finding. 
In addition, the parameters for the classifiers were not tuned. As such, this result does not allow for conclusions on the best approach for prediction.
The embeddings achieved the same median AUC of 0.675 across tasks as the models using count-based features.
Further summary statistics are shown in Table\,\ref{tab:predictive_performance_by_task}, and all task specific metrics are available in the supplementary material (Table S2).

\subsubsection{Interpretability of embeddings: comparison to ICD-10 categories}\label{icd_10_rediscovery}
\begin{figure}[h]
\includegraphics[width=1\textwidth]{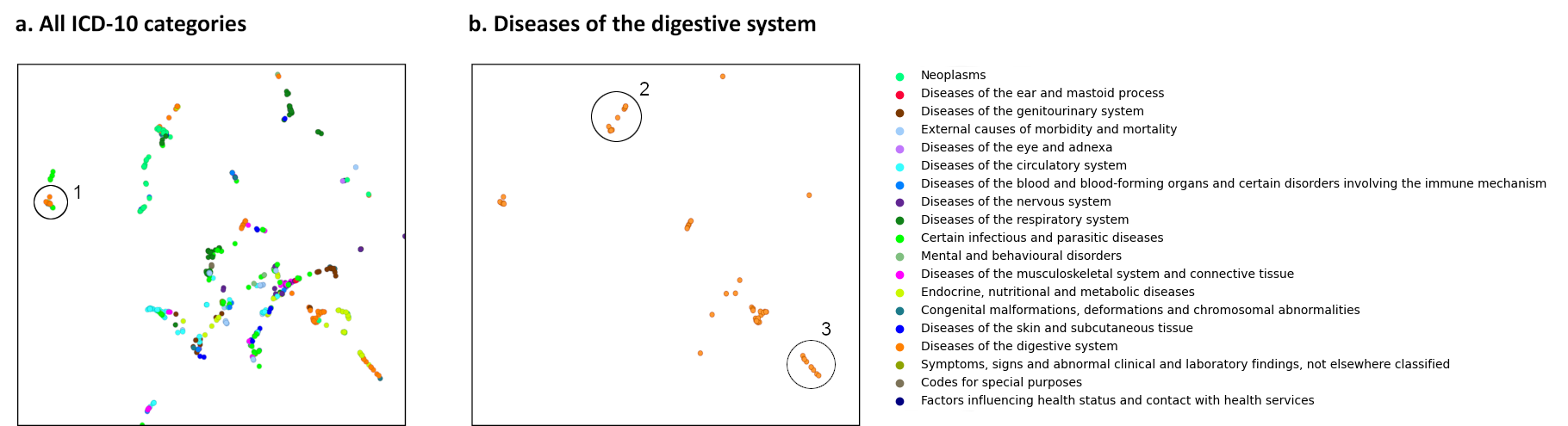}
\centering
\caption{Diagnosis embeddings mapped to two dimensions using UMAP and color-coded by the ICD-10 category to which they belong. Left: Plot shows all diagnosis features. Region 1: Alongside the grouping of hepatobiliary diseases, from diseases of the digestive system, portal vein thrombosis and oesophageal
varices, from the diseases of the circulatory system category map closely. These circulatory diseases are often diagnosed as complications of  hepatobiliary diseases, such as chronic liver diseases. Right: Plot
filtered to show only diagnosis features from diseases of the digestive system ICD-10 category. Region 2: demonstrates a subcategory
of diagnoses relating to oral and salivary gland disorders, e.g., stomatitis, oral cysts, diseases
of the tongue. Region 3: demonstrates a subcategory of diagnoses relating to intestine disorders, e.g., intestinal malabsorption, vascular disorders of the intestine.}
\label{fig:embeddings}
\end{figure}

To assess clinical interpretability, we visually examined how representations of diagnosis embeddings mapped on a two-dimensional space using Uniform Manifold Approximation and Projection (UMAP). 
Each diagnosis was colored according to the ICD-10 category to which it belongs. 
We found that the generated representations broadly captured clinically accepted similarities, as shown by diagnoses of the same categories forming groups together (see Figure\,\ref{fig:embeddings}). 
When examined closely, groups of the same category map to different regions of the 2D space represent sub-categories within a broader ICD-10 category. 
For example, in diseases of the digestive system, there is a group relating to diseases of the oral cavity and salivary gland in one region and hepatobiliary diseases in another (Figure \ref{fig:embeddings}). 
In addition, we see that the embeddings are able to capture phenotypic similarities that are not identified by the ICD-10 system, as shown by diagnosis embeddings of related diseases from different categories mapping closely together (e.g., see Figure \ref{fig:embeddings}, Region 1).

\section{Discussion}\label{discussion}
Selection of indications for a new therapeutic MoA remains a critical question in drug development. The depth and diversity of RWD is increasingly being explored to improve understanding of disease biology and has applications across the entire product lifecycle.
The presented work shows that representation learning applied to RWD is able to identify and rank diagnoses suited for a given MoA. 

This paper applies the indication finding approach to an established MoA with multiple approved indications and positive and negative results in clinical trials for referential definition and validation. 
The approach can also be applied to novel MoAs without approved indications or validation in clinical trials. 
In such cases the selection of referentials is based on evidence from a wider range of sources including -omics, literature, and knowledge graphs; referentials would be defined as diseases with the strongest association to the target on which the novel MoA acts. Over time, as a broader evidence base is established in relation to the MoA of interest, the diseases selected as referentials can be refined/updated to re-generate a ranked list based on the latest evidence available.

Leveraging a broad feature space is an important step in the analysis to characterise the patient phenotype and enable a wide search. 
For the anti-IL-17A case study we defined diagnosis, prescription and diagnostic procedure features, including whether a specific lab test was ordered. 
If available, lab test results could be incorporated by recording high/low/normal or normal/abnormal outcomes.
Since lab tests form an important component in some diagnoses, this may be desired by clinicians. 
However, many RWD provide no (or only limited) access to the results of lab tests.  
It may be argued that if test results were clinically significant for a patient they would likely lead to a change in the patient treatment journey which would be detectable across the other available features.

The overall ranking of each indication is an aggregation of individual referential proximity results. 
A diagnosis which is similar to all referential diseases (and therefore ranks highly) is more likely to share the same underlying pathway, related to its pathogenesis, as the referentials. Whilst in the case study, we had equal confidence in the efficacy of our MoA across referentials and hence the importance of the IL-17A pathway to them, in cases where confidence varies across referentials this could be incorporated by weighting in the final aggregation. Moreover, different methods can be used for aggregating the results across referentials. In this work, the median rank was used, but direct summing of the proximity scores across the referentials is also an effective aggregation method.

The method is able to take advantage of large patient cohorts (e.g., 17.8 million in the IL-17A case study). 
Although it is beneficial to have a broad cohort to search a large space and maximise the evidence used, other RWD sources may lead to smaller cohorts. 
We performed an ablation study to quantify the impact of the cohort size on the results. We found that reducing cohort size by up to a factor of three left positive validations within the top 100 ranked diagnoses unchanged, whereas reducing cohort size by a factor of six resulted in a reduction of 33\% in the positive validations ranked in the top 100 indications for some runs. 
This indicates that the approach is also successful with smaller cohort sizes.

The approach for indication finding described in this paper is not tied to a specific method for deriving clinical event representations. 
In future work, the exploration of embeddings derived from language models that have been adapted to the domain of EHR is a promising direction, e.g. BEHRT \cite{li_behrt_2020}. 
Embeddings derived with these types of models have also proven to robustly perform in clinical tasks, such as readmission prediction \cite{jiang_health_2023} or mortality prediction \cite{rupp_exbehrt_2023}.

In conclusion, we have introduced a data-driven, quantitative approach to indication finding, and a systematic framework for evaluation. We demonstrated successful application of this approach for anti-IL-17A. This approach can be leveraged, alongside traditional methods of indication finding, to increase the likelihood of success in clinical trials.

\section{Methods}\label{methods}

\subsection{Set up of approach}\label{set_up_of_approach}

We selected anti-IL-17A as the MoA to be investigated to enable simple quantitative evaluation of the approach.
Anti-IL-17A is an antibody-based therapy that antagonizes interleukin 17A, an inflammatory signaling protein affecting multiple downstream processes. 
There are a number of approved indications, as well as several Phase 2 or Phase 3 trials that have been completed with varying outcomes, providing the validations for quantitative evaluation

\begin{figure}[h]
\includegraphics[width=1\textwidth]{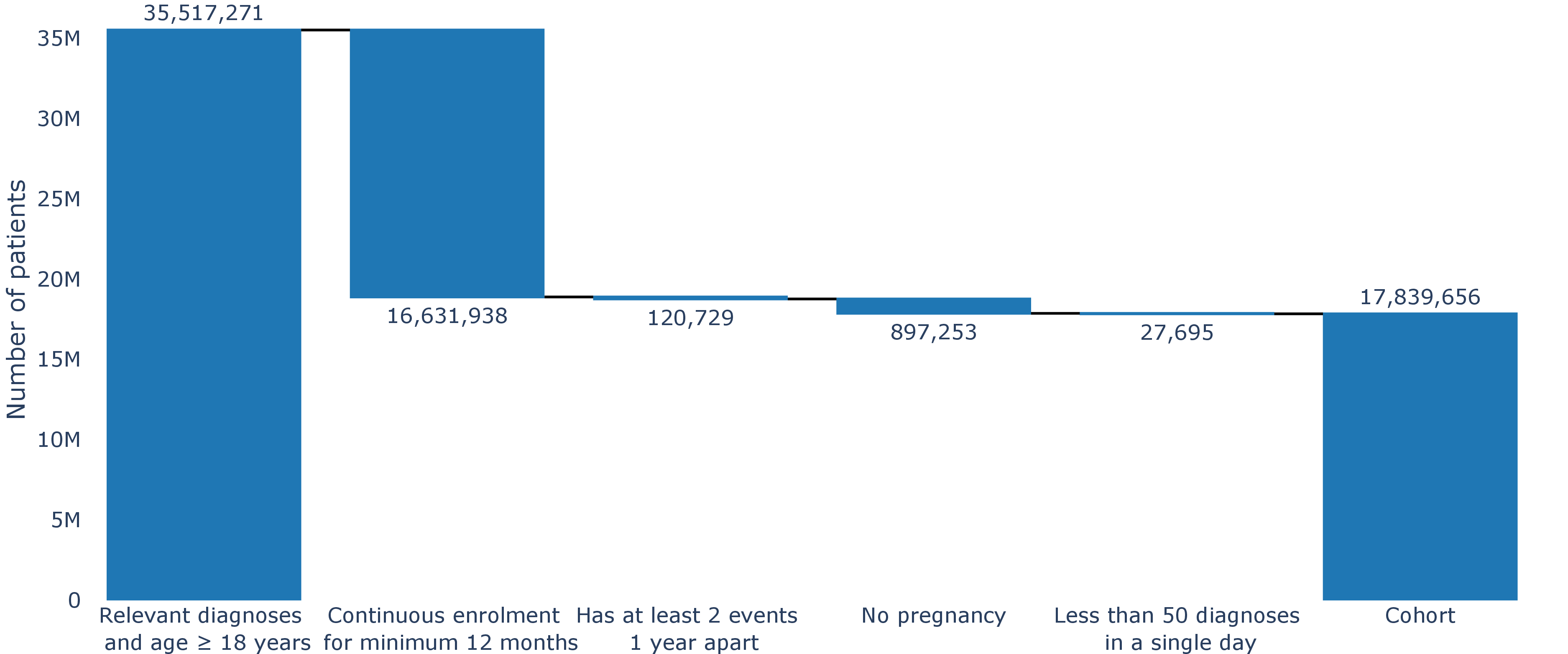}
\centering
\caption{Waterfall to illustrate the impact of inclusion / exclusion criteria on cohort size.}
\label{fig:waterfall}
\end{figure}

We used Komodo Health's comprehensive dataset \cite{komodo_reference} (see \nameref{disclaimer}) to generate a cohort of 17,839,656 patients based on a number of inclusion and exclusion criteria.
The observation window was January 2018 to December 2022, with over 90\% of patients observed for at least 3 years.
The basis for the cohort definition was the list of diseases most associated with the IL-17A based on Open Targets review \cite{ochoa_next-generation_2023}.
Diseases identified were removed from this list if they were too broad (i.e., terms that could refer to multiple distinct diseases such as ‘cancer’ or ‘immunodeficiency’), associated with the wrong directionality (i.e. low IL-17A) or related to cancer, as oncological diagnoses could not be accurately defined with the Komodo dataset (e.g., because histology results were not available). 
After these filters, the 17 diseases with the highest overall association were chosen as clinical inclusion criteria. 
Table \ref{table: cohort_inclusion_diseases} specifies the disease names and parent ICD-10 codes used for cohort inclusion criteria. 
In addition, patients were required to be over the age of 18, to have had a continuous period of enrollment with an insurer for a minimum of 12 months, and to have had two medical visits at least one year apart between January 2018 and December 2022. 
These latter two conditions helped to ensure that patients in the cohort had sufficient interaction with the healthcare system. 
Patients that had a pregnancy-related diagnosis were excluded from analysis (ICD-10 codes O00-O99 for pregnancy and P00-P96 for conditions originating in the perinatal period, respectively), due to the different physiology from non-pregnant patients. 
Patients that had more than 50 diagnoses in a single day recorded in the data were also excluded due to data quality concerns. See Figure \ref{fig:waterfall} for the cohort waterfall.

\subsubsection{Feature definition}\label{feature_definition}
We generated 1,519 diagnosis, 1,733 prescription, 114 procedure and 8 demographics features. Each feature is linked to a relevant set of codified events within clinical records.
Definitions ensure comprehensive coverage of the patient journey as well as granular analysis of key events.

For diagnoses, all ICD-10 codes (version: 2019) with the same starting three-characters were grouped.
This approach was taken for all ICD-10 codes except those in chapter XVIII (symptoms, signs and abnormal clinical and laboratory findings, not elsewhere classified), chapter XIX (injury poisoning and consequences of external causes) and chapter XXI (factors influencing health status and contact with health services). 
Codes from category XVIII were excluded because they may not be consistently documented in health insurance data as claims are not usually raised against them. 
Codes from categories XIX and XXI were excluded because they do not provide information relating to relevant disease pathology for our analysis. 
We also defined granular diagnosis features for those diseases most likely to be effectively treated by the selected MoA. 
All ICD-10 codes linked to the 50 diseases most closely associated with IL-17A according to Open Targets \cite{ochoa_next-generation_2023} were reviewed and grouped by a clinician according to clinical phenotype (Section\,\ref{set_up_of_approach}).
For example, asthma diagnosis codes were grouped into three granular features: mild, moderate and severe asthma.

For prescriptions, all NDC codes related to a given class in the First Databank, Inc.'s Enhanced Therapeutic Classification System (see \nameref{disclaimer}) were grouped to create 1909 therapeutic class features, these classes represent biologically distinct groups of medications \cite{fdb_reference}.

For diagnostic procedures, all CPT and HCPCS codes related to a CCS class of diagnostic procedure in the ICD-10-PCS classification system were grouped \cite{ccs_reference} with clinical review of the list to remove categories of procedures not related to diagnostics.

Grouping features ensures comprehensive coverage of the patient journey whilst managing computational cost and high overlap between features. 
However, broad feature categories reduce the ability for granular differentiation that may be important for certain areas.
In particular, the CCS for ICD-10-PCS level 2 classification system defines very wide-ranging groups of diagnostic procedures, e.g., organ panel blood tests (related to multiple organ systems, including liver function tests and renal profile).

To mitigate for this, we also created a small number of granular prescription and procedure features linked to individual relevant codes or small groups of codes that represent procedures/prescriptions used in the management of referential diseases, e.g., skin biopsy for psoriasis.
This selection was based on a review of the clinical guidelines \cite{ward_2019_2019,singh_2018_2019,elmets_joint_2021,menter_joint_2020,elmets_joint_2019,menter_joint_2019,elmets_joint_2019-1}. 

The demographics features used were gender (male/female), and age buckets 18-27, 28-37, 38-47, 48-57, 58-67, 67+. 

ICD-10 codes linked to diagnoses that the studied MoA is approved to treat were defined as referential features (\emph{plaque psoriasis}, \emph{psoriatic arthritis}, \emph{anklyosing spondylitis}) and are displayed in Table \ref{table: ref_and_control_icd_codes}. 
\emph{Non-radiographic ankylosing spondylitis} (an additional approved indication) was grouped with the broader indication \emph{ankylosing spondylitis}. 
Approved pediatric indications (\emph{juvenile psoriatic arthropathy}, \emph{enthesitis-related JIA}) were not used because our analysis was performed on an adult cohort. 
Features for diagnoses that the MoA has been shown to be effective or ineffective in treating during phase II/III clinical trials were linked to relevant ICD-10 codes for use as positive and negative validations, respectively, see Table \ref{table: ref_and_control_icd_codes}.

\newcolumntype{L}[1]{>{\raggedright\arraybackslash}p{#1}}

\begin{table}
\centering
\begin{tabular}{llL{4cm}L{1.7cm}}
\toprule
\textbf{Feature name }            & \textbf{Feature type}     & \textbf{ICD 10 codes} & \textbf{Trial ID } \\
\midrule
Plaque psoriasis     & Referential & L400             &          NCT01365455, NCT05232871  \\
Psoriatic arthritis     & Referential &L405, M070, M071, M072, M073             & NCT01752634, NCT05232872    \\
Anklyosing spondylitis     & Referential &M450, M451, M452, M453, M454, M455, M456, M457, M458, M459, M45A        & NCT01649375, NCT05232873           \\
Rheumatoid arthritis     & Positive validation &M05, M06              & NCT01350804          \\
Rosacea                  & Positive validation &L71              & NCT03079531          \\
Hidradenitis suppurativa & Positive validation & L732             & NCT03713619          \\
Non-infectious uveitis       & Positive validation & H2000, H2001, H2002, H2004             & NCT00685399          \\
Giant cell arteritis     & Positive validation &M315, M316, G737       & NCT03765788          \\
Crohn’s disease     & Negative validation & K50             &NCT01009281           \\
Congenital ichthyosis,       & Negative validation & Q80            &NCT03041038           \\
Atopic dermatitis       & Negative validation &   L20           &NCT02594098           \\
Bullous pemphigoid       & Negative validation &  L120            & NCT03099538          \\
COVID-19           & Negative validation  & U071, U10             &NCT04724629  \\ 
 \bottomrule
\end{tabular}
\caption{ICD codes of positive validations, negative validations and referentials, along with their NCT trial ID.}\label{table: ref_and_control_icd_codes}
\end{table}

\subsection{Embedding approach}\label{embedding_approach}

SPPMI was implemented as described by Levi and Goldberg \cite{levy_neural_2014} apart from  two changes:
firstly, the context window was defined based on calendar dates instead of clinical events to account for the fact that within a given patient visit to a clinician or care provider, the order in which codes appear on their record does not have temporal significance. Secondly, the demographic features gender and age were injected in the co-occurrence matrix calculation by ensuring that these features appeared exactly once in every analyzed time window.

During the exploration of hyperparameters, we varied the window size $w$ in days ($180$, $360$) for computing the co-occurrence matrix, as well as the embedding dimensions $d$ ($25$, $50$, $75$, $100$), the smoothing parameter $\alpha$ ($0.25$, $0.5$, $0.75$, $1$) and the eigenvalue weighting $v$ ($0$, $0.25$, $0.5$, $0.75$, $1$). 
The PPMI matrix was not shifted. These experiments were carried out on the entire dataset.
The highest scoring set of parameters was $w= 360/180$, $d=75$, $\alpha=0.75$, $v=0.75$. 
Parameters with the 360 day window were used to generate the results outlined in this paper.
The parameter choice had limited effect on the results with all sensible parameters ranking more positive validations high than negative ones. 
The complete hyperparameter exploration is provided in the supplementary material (Table S1). 
One run was performed for each parameter setting and no additional stability runs were performed (see Section\,\ref{ranked_list_of_indications}). 
Stability of the highest indications was only assessed for the best run used to obtain the final results.

The best model was selected based on its quantitative indication finding performance.
To that end, a scoring function sensitive to the ranking of positive and negative validations was defined.
Each positive validation received a score between 0 and 5 depending on their rank, with 5 points awarded for a rank in the top 30.
Negative validations received a score between -0.5 and 0 depending on their rank, with a penalty of -0.5 if the indication were to rank in the top 30. 
See Table \ref{tab:scoring_function} for details.
Scores were summed across validations to arrive at the overall indication finding score used in the evaluation.

\begin{table}[]
\begin{tabular}{lp{4cm}p{4cm}}
\toprule
 \textbf{Ranking of validation} & \textbf{\shortstack{Score assigned to positive \\validations based on rank}} & \textbf{\shortstack{Score assigned to negative \\validations based on rank}}   \\
 \midrule
1-30 &  5& -0.5    \\
 31-60& 4 & -0.4    \\
61-90 & 3 &  -0.3  \\
91-120 &2 &  -0.2  \\
121-150 & 1 &  -0.1  \\
\botrule
\end{tabular}
\caption{Score contribution for positive or negative validations, based on their ranking in a given section of the list. The sum of scores across validations was used for quantitative assessment of indication finding performance.}
\label{tab:scoring_function}
\end{table}

\subsection{Generation of ranked list of indications}\label{ranked_list_of_indications}
We ranked indications that were sufficiently specific for potential investigation in clinical trials. 
Therefore all diagnosis features relating to ICD-10 classes containing `other', `unspecified' or `in diseases classified elsewhere' in their name were not included in the ranking process. 
In addition, diseases related to `external causes of morbidity and mortality' (from chapter XX of the ICD-10 system) were excluded from ranking.

The proximity between each diagnosis-referential pair was calculated using the cosine similarity between their representations. 
Indications were ranked with respect to each referential by their proximity score. The median rank across referentials was calculated for each indication and used to order the list for the rankings.

To determine the stability of an indication's position on the ranked list with regard to the selection of patients available in the data, we re-ran the pipeline five times on different random samples of 50\% of the patient cohort.
From this, we obtained a distribution of the proximity scores for each diagnosis-referential pair; the lower bound of the 95\,\% percentile interval of each distribution was calculated using a normal distribution fitted to the sample runs. 
These lower bound proximity scores were then used to re-rank indications for each referential. 
The delta between the ranks obtained using the full data set and the ranks obtained using the lower bound scores were calculated, for each indication-referential pair. The median delta across each referential is subsequently determined, for each indication, in order to give equal weight to all referentials and reduce sensitivity to extreme ranks. Since we are considering the top 200 diagnoses, diagnoses having a median delta of over 200 for at least one referential are filtered out as `unstable'.
Following the removal of `unstable' diagnoses, the final ranked list is obtained.

\section*{Data availability}\label{data_availability}
The data that support the findings of this study are available from Komodo. Restrictions and licensing fees apply.

\backmatter

\bibliography{sn-bibliography.bib}

\section*{Acknowledgments}\label{acknowledgements}
We would like to recognize the contribution of several others in reviewing and guiding the development of this pipeline. Namely, Aalok Parkash, Kirtikumar Shinde, Ismail Ahmady, L\'eon Jouet, Hadeel Mustafa, Muhammad Anwar, Julian Waton, Lise Diagne, Giulio Morina, Kevin Riera, Steven Ler, Leigh Jansen, Joachim Bleys, and Meredith Langstaff.

\section*{Author Contributions}\label{author_contributions}
Project management: V.S.; Methodology: M.E, C.A., A.A.; Data processing and analytics: A.A.; Writing: V.S., E.B., A.A., M.E; Approval of final manuscript: all authors.

\section*{Competing interests}\label{competing_interests}
All authors are employed at McKinsey and Company at the time of writing this article. The approach defined in this article may be deployed by McKinsey and Company during client projects.

\section*{Disclaimers}\label{disclaimer}
Komodo Health, Inc. makes no representation or warranty as to the accuracy or completeness of the data (“Komodo Materials”) set forth herein and shall have, and accept, no liability of any kind, whether in contract, tort (including negligence) or otherwise, to any third party arising from or related to use of the Komodo Materials by McKinsey and Company. Any use which McKinsey and Company or a third party makes of the Komodo Materials, or any reliance on it, or decisions to be made based on it, are the sole responsibilities of McKinsey and such third party. In no way shall any data appearing in the Komodo Materials amount to any form of prediction of future events or circumstances and no such reliance may be inferred or implied.

The FDB Enhanced Therapeutic Classification System\textsuperscript{TM} is a proprietary advanced drug classification system with the FDB MedKnowledge\textsuperscript{\textregistered} database that provides multiple ways to categorize drugs for easy formulary maintenance and drug selection. It allows drugs to reside in multiple therapeutic classes, with links to drug concepts at any level of the hierarchy. Reprinted with permission by First Databank, Inc. All rights reserved. \textcopyright 2023

\newpage
\appendix
\section*{Appendix}
\counterwithin{figure}{section}

\begin{table}[h]

\begin{tabular}{@{}l@{\hspace{1cm}}l@{}}
\toprule
\textbf{Disease name} & \textbf{ICD-10 code} \\
\midrule
Psoriasis &L40\\
Psoriatic arthritis &L405, M070, M071, M072, M073\\
Ankylosing spondylitis & M45\\
Psoriasis vulgaris & L400\\
Spondyloarthropathy & M461, M468, M469\\
Rheumatoid arthritis & M05, M06\\
Hidradenitis suppurativa & L732\\
Uveitis & H200, H201, H208, H209\\
Pustular psoriasis & L401\\
Ichthyosis & Q80\\
Lupus nephritis & N085\\
Temporal arteritis &M30, M31\\
Behcet's syndrome &M352, N778\\
Grave's ophthalmopathy &E0501, H588\\
Asthma & J453, J454, J455\\
Systemic lupus erythematosus &M32\\
Chronic obstructive pulmonary disease &J44\\
\botrule
\end{tabular}
\caption{Diseases used as cohort inclusion criteria along with ICD-10 parent codes. Patients with more granular ICD-10 codes were also included.}\label{table: cohort_inclusion_diseases}%
\end{table}

\begin{table}[h]
\begin{tabular}{llll}
\toprule
 \textbf{Target} & \textbf{Prevalence} & \textbf{Count-based} & \textbf{SPPMI}  \\
 \midrule
\text{Stroke} & 6.11\% & \textbf{0.67} &\textbf{0.67} \\
\text{Rheumatoid Arthritis} & 11.19\% & 0.64 &\textbf{0.65} \\
\text{Neutropenia} & 1.44\% & 0.69 &\textbf{0.70}  \\
\text{Heart attack} & 5.42\% & \textbf{0.68} & \textbf{0.68}  \\
\text{Type II diabetes mellitus} & 17.86\% & \textbf{0.65} & 0.64\\
\text{CKD} & 12.80\% & \textbf{0.70} & 0.69\\
\botrule

\end{tabular}
\caption{AUC scores in the predictive task broken down by method and disease, along with prevalence of the disease. Highest values per row are marked in bold. For every model the AUC with the best classifier is shown.}
\label{tab:predictive_performance_by_task}
\end{table}

\end{document}